# Partial differential equation system for binarization of degraded document images

Youjin Liu, Yu Wang[*]

College of Mathematics and Statistics, Chongqing University, Chongqing 401331

**Abstract**

In recent years, partial differential equation (PDE) systems have been successfully applied to the binarization of text images, achieving promising results. Inspired by the DH model and incorporating a novel image modeling approach, this study proposes a new weakly coupled PDE system for degraded text image binarization. In this system, the first equation is designed to estimate the background component, incorporating both diffusion and fidelity terms. The second equation estimates the foreground component and includes diffusion, fidelity, and binarization source terms. The final binarization result is obtained by applying a hard projection to the estimated foreground component. Experimental results on 86 degraded text images demonstrate that the proposed model exhibits significant advantages in handling degraded text images.

---

[*] Corresponding author
  *Email address:* 202206021015@stu.cqu.edu.cn (Youjin Liu), wangyu_cqu2023@163.com (Yu Wang)

# 1 Introduction

The quality of the binarization results for text images has a decisive impact on subsequent stages such as feature extraction and recognition processing in Optical Character Recognition (OCR) systems, and it may even affect the entire process and performance of Document Analysis and Recognition (DAR) technologies. Because of this, text image binarization has become a key focus of widespread attention and in-depth research in academia. Scholars are dedicated to improving the accuracy and efficiency of binarization processing to ensure the stable operation of OCR systems and the overall performance of DAR technologies.

A wide range of text image binarization methods have been proposed, which can generally be categorized into four main types: clustering-based methods [1][2][3], deep learning-based methods[4]-[12], thresholding-based methods[13]-[26], and partial differential equation (PDE)-based methods[27]-[36][39][40].

In recent years, many researchers have introduced PDEs into text image binarization studies, achieving outstanding processing results while also providing theoretical explanations. This integration of PDEs with image processing has become a trending research direction with promising future developments. Recently, some researchers have applied PDE systems to text image binarization by designing different PDE formulations to achieve effective binarization.

In [30], Guo et al. proposed a coupled PDE system for degraded text images with low contrast or contrast variations. This system combines fourth-order indirect diffusion, an adaptive shock filter operator, and a time-dependent binary classifier with phase separation properties. In [39], Jacobs and Celik constructed a nonlinear PDE system in which the main equation is responsible for binarization, while the auxiliary equation controls the evolution of key threshold parameters within the main equation. In [40], Du et al. introduced a nonlinear diffusion equation with a dynamic threshold function as a source term. This dynamic threshold function is governed by another evolving PDE. In [36], Du and He proposed a PDE system consisting of two nonlinear diffusion equations that can simultaneously perform restoration and binarization of degraded text images. One equation is responsible for text restoration, and the other is used for text binarization. The DH model, in the field of PDE-based text binarization, was the first to apply the idea of image decomposition, breaking the text image into the product of foreground and background components. The two evolutionary equations in the DH model can simultaneously estimate the foreground and background components of the text image. Experimental results show that the model performs well in binarizing degraded text images with five typical degradation types: transparency, complex backgrounds, page stains, stamps, and ink smudges.

Based on the above research on PDE systems in the field of text binarization, this paper,

building upon the DH model and incorporating new image modeling, proposes a new weakly coupled PDE system for binarizing degraded text images. The first equation is used to estimate the background component, including diffusion and fidelity terms, while the second equation is used to estimate the foreground component, containing diffusion, fidelity, and binarization source terms. The final binarized result is obtained by performing a hard projection on the estimated foreground component.

The main contributions of this work based on the DH model are two-folds:

(1) A new image generation model is proposed to describe degraded document images, which decomposes the degraded document image into the sum of background and foreground components.

(2) A new weakly coupled partial differential equation system model is proposed to achieve binarization of degraded document images.

The paper is divided into five sections. Section 2 introduces the DH model and our image generation model, followed by the presentation and analysis of our weakly coupled partial differential equation system model. Section 3 discusses the numerical solution of the model. Section 4 presents the experimental results and evaluation of the method. The final section provides the conclusion.

**2 Proposed model**

In this section, we first introduce the DH model, followed by the degraded text image generation model in our model. Additionally, we propose a new weakly coupled partial differential equation system model and analyze the role of each term, which is ultimately used to achieve binarization of degraded document images.

**2.1 The DH model**

DH model based background estimation technology defines the degraded text image s(x) on the continuous region $\Omega \subset \mathcal{R}^2$ as follows:

$$s(\mathrm{x}) = b(\mathrm{x})u(\mathrm{x}) + n(\mathrm{x}), \ \mathrm{x} \in \Omega \tag{1}$$

where $b(\mathrm{x})$ represents the background image, and $u(\mathrm{x})$ represents the foreground image, which is used to represent the text; $n(\mathrm{x})$ represents the noise image, used to indicate the difference between the text image and the background and foreground images.

Combining the image generation model in (1), the DH model propose a weakly coupled nonlinear PDE system, as expressed below:

$$\begin{cases} \dfrac{\partial b}{\partial t} = \lambda_{11} \cdot \Delta b + \lambda_{12} \cdot u(s-bu), & \text{in } \Omega \times (0,+\infty) \\ \dfrac{\partial u}{\partial t} = \lambda_{21} \cdot div\left(g_0\left(|\nabla u|^2\right)\nabla u\right) + \lambda_{22} \cdot b(s-bu) + B_0(s,u), & \text{in } \Omega \times (0,+\infty) \end{cases} \quad (2)$$

with the initial conditions:

$$b(\mathrm{x},0) = 1, \; u(\mathrm{x},0) = s(\mathrm{x}), \; \mathrm{x} \in \Omega \quad (3)$$

In the above, $\lambda_{11}, \lambda_{12}, \lambda_{21}, \lambda_{22}$ are all non-negative constants that control the strength of the corresponding terms, and $u(\mathrm{x},t)$ is the foreground component at time $t>0$, $g_0$ is the diffusion coefficient:

$$g_0\left(|\nabla u|^2\right) = \dfrac{1}{1+|\nabla u|^2/\kappa}, \; \kappa = mean\left(|\nabla u(\mathrm{x},t)|^2\right) \quad (4)$$

The binarization source $B_0(s,u)$ in (2) is defined as:

$$\begin{aligned} B_0(s,u) &= \lambda_{23}\omega(\mathrm{x})u(1-u)(u-c(\mathrm{x})) \\ &\quad + (1-\lambda_{23})(1-\omega(\mathrm{x}))\mu(t)u(1-u)(u-s_{\min}) \end{aligned} \quad (5)$$

where $0 < \lambda_{23} < 1$ is a constant, and $s_{\min} = \min\limits_{\mathrm{x} \in \Omega}\{s(\mathrm{x})\}$, meaning the minimum value of the image $s(\mathrm{x})$ within the image region $\Omega$, $\mu(t) = 1 - e^{-t/20}$.

In (5), $\omega(\mathrm{x})$ is a weighting function, and $c(\mathrm{x})$ is a local function defined as follows:

$$\omega(\mathrm{x}) = \dfrac{d(\mathrm{x}) - \min\limits_{\mathrm{x} \in \Omega} d(\mathrm{x})}{\max\limits_{\mathrm{x} \in \Omega} d(\mathrm{x}) - \min\limits_{\mathrm{x} \in \Omega} d(\mathrm{x})}, \; \mathrm{x} \in \Omega \quad (6)$$

$$c(\mathrm{x}) = m_\mathcal{B}(\mathrm{x})\bar{s}_\mathcal{F}(\mathrm{x}) + m_\mathcal{F}(\mathrm{x})\bar{s}_\mathcal{B}(\mathrm{x}), \; \mathrm{x} \in \Omega \quad (7)$$

where the function $d(\mathrm{x})$ is defined as:

$$d(\mathrm{x}) = \log(1 + |\bar{s}_\mathcal{B}(\mathrm{x}) - \bar{s}_\mathcal{F}(\mathrm{x})|), \; \mathrm{x} \in \Omega \quad (8)$$

Next, we further explain the local fuzzy cluster method for defining the components $m_\mathcal{F}(\mathrm{x})$, $m_\mathcal{B}(\mathrm{x})$, $\bar{s}_\mathcal{F}(\mathrm{x})$ and $\bar{s}_\mathcal{B}(\mathrm{x})$.

For $\forall \mathrm{x} \in \Omega$, define $\mathcal{O}_\rho(\mathrm{x}) = \{\mathrm{y} \in \mathcal{R}^2 : |\mathrm{y} - \mathrm{x}| < \rho\}$ as the local region centered at $\mathrm{x}$, where $\rho > 0$ is the radius of the local neighborhood. The weighted mean of $s(\mathrm{x})$ in $\mathcal{O}_\rho(\mathrm{x})$ is defined as:

$$\bar{s}_\rho(\mathrm{x}) = \int_{\mathcal{O}_\rho(\mathrm{x})} K_\rho(\mathrm{x}-\mathrm{y})s(\mathrm{y})d\mathrm{y} = \int_{\mathcal{R}^2} K_\rho(\mathrm{x}-\mathrm{y})s(\mathrm{y})d\mathrm{y} \quad (9)$$

where $K_\rho$ is the mollifying kernel (mollifying function):

$$K_\rho(h) = \begin{cases} C\rho^{-2}\exp(-1/(1-|h|^2/\rho^2)), & \text{if } |h| < \rho \\ 0, & \text{if } |h| \geq \rho \end{cases} \quad (10)$$

where $C > 0$ ensures $\int_{\mathcal{R}^2} K_\rho(h)dh = 1$. The mollifier has compact support. The region $\Omega_1$ is composed of both text and background points, while the region $\Omega_2$ consists only of background points. This region $\Omega$ can be split as $\Omega = \Omega_1 \cup \Omega_2$, and $\Omega_1 \cap \Omega_2 = \Phi$, where $\Omega_1$ represents the area containing text, and $\Omega_2$ represents the area containing only background.

Then,
$$|s(\mathrm{x}) - \bar{s}_\rho(\mathrm{x})| \gg 0, \ \mathrm{x} \in \Omega_1 \tag{11}$$
and
$$|s(\mathrm{x}) - \bar{s}_\rho(\mathrm{x})| \approx 0, \ \mathrm{x} \in \Omega_2 \tag{12}$$

Thus, based on (11) and (12), two fuzzy member functions $m_{\mathcal{F}}(\mathrm{x})$ and $m_{\mathcal{B}}(\mathrm{x})$ are defined as:

$$\begin{aligned} m_{\mathcal{F}}(\mathrm{x}) &= \frac{1}{2} - \frac{1}{2}\tanh((s(\mathrm{x}) - \bar{s}_\rho(\mathrm{x}))/\varepsilon) \\ m_{\mathcal{B}}(\mathrm{x}) &= \frac{1}{2} + \frac{1}{2}\tanh((s(\mathrm{x}) - \bar{s}_\rho(\mathrm{x}))/\varepsilon) \end{aligned} \tag{13}$$

in this case, the constant $\varepsilon > 0$ is used to control the gradient between $\varepsilon > 0$ in the transition region. $\mathcal{F}/\mathcal{B}$ represents the foreground (or background) of the input image $s$, and the value of $m_{\mathcal{F}}(\mathrm{x})/m_{\mathcal{B}}(\mathrm{x})$ corresponds to the degree of belonging to the foreground $\mathcal{F}$ (or background $\mathcal{B}$).

Let $\mathcal{F}_{\mathrm{x}}/\mathcal{B}_{\mathrm{x}}$ be the local foreground / background clusters of the image $s$ in the neighborhood of $\mathcal{O}_\rho(\mathrm{x})$, defined as the local foreground / background cluster centers in the following way:

$$\begin{aligned} \bar{s}_{\mathcal{F}}(\mathrm{x}) &= \frac{\int_{\mathcal{O}_\rho(\mathrm{x})} K_\rho(\mathrm{x}-\mathrm{y}) m_{\mathcal{F}}(\mathrm{y}) s(\mathrm{y}) d\mathrm{y}}{\int_{\mathcal{O}_\rho(\mathrm{x})} K_\rho(\mathrm{x}-\mathrm{y}) m_{\mathcal{F}}(\mathrm{y}) d\mathrm{y}} = \frac{\int_{\mathcal{R}^2} K_\rho(\mathrm{x}-\mathrm{y}) m_{\mathcal{F}}(\mathrm{y}) s(\mathrm{y}) d\mathrm{y}}{\int_{\mathcal{R}^2} K_\rho(\mathrm{x}-\mathrm{y}) m_{\mathcal{F}}(\mathrm{y}) d\mathrm{y}} \\ \bar{s}_{\mathcal{B}}(\mathrm{x}) &= \frac{\int_{\mathcal{O}_\rho(\mathrm{x})} K_\rho(\mathrm{x}-\mathrm{y}) m_{\mathcal{B}}(\mathrm{y}) s(\mathrm{y}) d\mathrm{y}}{\int_{\mathcal{O}_\rho(\mathrm{x})} K_\rho(\mathrm{x}-\mathrm{y}) m_{\mathcal{B}}(\mathrm{y}) d\mathrm{y}} = \frac{\int_{\mathcal{R}^2} K_\rho(\mathrm{x}-\mathrm{y}) m_{\mathcal{B}}(\mathrm{y}) s(\mathrm{y}) d\mathrm{y}}{\int_{\mathcal{R}^2} K_\rho(\mathrm{x}-\mathrm{y}) m_{\mathcal{B}}(\mathrm{y}) d\mathrm{y}} \end{aligned} \tag{14}$$

**2.2 Document image formation of our model**

The main challenge in binarizing text images is the presence of multiple degradation factors, such as uneven lighting, background noise, and low contrast. Among these, background estimation [41]-[44], as a commonly used preprocessing technique, can effectively reduce the complexity of subsequent binarization tasks.

The binarization method based on background estimation first estimates the background of the original text image, and then uses this background image to obtain an easier binarized text image. One common background compensation method is division[41][42]:

$$\hat{s}(i,j) = \beta \cdot s(i,j)/b(i,j) \quad (15)$$

where $\hat{s}(i,j)$ is the preprocessed version of the observed image $s(i,j)$, and $b(i,j)$ is the estimated background image. However, based on the division compensation method, when $b(i,j)$ is dark (or has low contrast), $1/b(i,j)$ becomes large, which may lead to overcompensation and a stronger influence of the background in certain areas. Another common background compensation method is simple subtraction [43][44]:

$$\hat{s}(i,j) = s(i,j) - b(i,j) \quad (16)$$

Based on the subtraction compensation method, it can be argued that this method improves the background compensation effect by subtracting the background. Thus, we adopt the modification to (16) as:

$$s(i,j) = \hat{s}(i,j) + b(i,j) \quad (17)$$

Inspired by (17), we propose an image model to describe degraded text images in continuous domains:

$$s(\mathrm{x}) = b(\mathrm{x}) + u(\mathrm{x}) + n(\mathrm{x}), \ \mathrm{x} \in \Omega \quad (18)$$

where $b(\mathrm{x})$ represents the background image, responsible for depicting the background; $u(\mathrm{x})$ represents the foreground image, used to depict the text; and $n(\mathrm{x})$ represents the error image, which accounts for the difference between the actual text image and the sum of the background and foreground images.

In the additive model of text image formation(18), we assume:

**(H1)** The background component $b(\mathrm{x})$ is a smooth function over the image domain $\Omega$.

**(H2)** The foreground component $u(\mathrm{x})$ is approximately a piecewise constant function over $\Omega$, where the values at text pixels are smaller than those at background pixels.

Fig. 1 provides an example illustrating an observed degraded text image with a non-uniform background under assumptions (H1) and (H2) in (18).

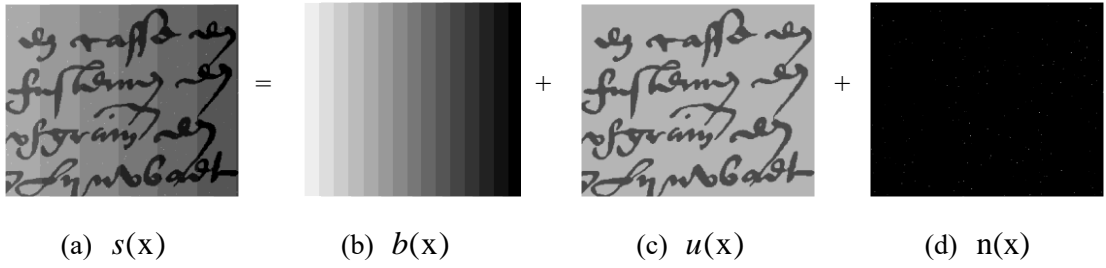

(a) $s(\mathrm{x})$      (b) $b(\mathrm{x})$      (c) $u(\mathrm{x})$      (d) $n(\mathrm{x})$

Fig.1 Illustration of image formation model

## 2.3 Proposed weakly coupled partial differential equation (PDE) model

Based on model (18) and assumptions (H1) and (H2), we propose a weakly coupled PDE model for the binarization of degraded text images:

$$\begin{cases} \dfrac{\partial b}{\partial t} = a_{11} \cdot \Delta b + a_{12} \cdot u(s-b-u), \text{ in } \Omega \times (0,+\infty) \\ \dfrac{\partial u}{\partial t} = a_{21} \cdot div\left(g\left(\left|\nabla^{\alpha} u\right|\right)\nabla u\right) + a_{22} \cdot b(s-b-u) + B(s,u), \text{ in } \Omega \times (0,+\infty) \end{cases} \quad (19)$$

with the initial conditions

$$b(\mathrm{x},0) = 1, u(\mathrm{x},0) = s(\mathrm{x}), \ \mathrm{x} \in \Omega \quad (20)$$

where $a_{i,j} > 0 \ (i,j=1,2)$ are parameters that control the strength of corresponding terms. $b(\mathrm{x},t)$ and $u(\mathrm{x},t)$ represent the background and foreground components at time $t > 0$, respectively. $B(s,u)$ is a binary classification term that distinguishes text pixels from background pixels, leading to the evolution of $u(\mathrm{x},t)$ over time and ultimately transforming it into a binarized image as $t \to \infty$.

Let $u(\mathrm{x},t^*)$ be the estimated foreground image at time $t^*$, then the binarized image $s_B(\mathrm{x})$ is obtained as:

$$s_B(\mathrm{x}) = \begin{cases} 1, \text{ if } u(\mathrm{x},t^*) \geq 0.5 \\ 0, \text{ if } u(\mathrm{x},t^*) < 0.5 \end{cases}, \mathrm{x} \in \Omega \quad (21)$$

Next, we briefly discuss the conservation, diffusion, and binarization properties of (19).

**(A) Conservation Term**

In the proposed model, the terms $u(s-b-u)$ and $b(s-b-u)$ are responsible for information exchange between the observed image and the background-foreground components. The term $(s-b-u)$ ensures that the sum of $b$ and $u$ evolves as a good approximation of the observed image.

**(B) Diffusion Term**

The first evolution equation in the PDE system (19) contains an isotropic diffusion term $\Delta b$, which ensures the smoothness of the background component $b$ during the iterative process.

Next, we analyze the nonlinear diffusion term in (19) in more detail. The diffusion term $div\left(g\left(\left|\nabla^\alpha u\right|\right)\nabla u\right)$ is responsible for the assumption that $u$ is a piecewise smooth function while preserving the text information in the input image $s$. When $\alpha=1$, this diffusion term corresponds to the nonlinear diffusion equation from model [37]:

$$\frac{\partial u}{\partial t} = div\left(g\left(\left|\nabla u\right|\right)\nabla u\right) \tag{22}$$

where the gradient magnitude $|\nabla u|$ is treated as an edge detector. The function $g(z)$ satisfies $g(z)\geq 0$ and decreases on the interval $[0,+\infty)$, ensuring $g(0)=1$ and $\lim_{z\to+\infty} g(z)=0$. In this paper, we choose $g(z)=\exp(-z^2/\varsigma^2)$, where $\varsigma = mean(z)$.

It is well known that fractional derivatives (FDs) are more suitable for describing complex structures in images compared to integer-order derivatives. There are multiple definitions of fractional derivatives, and the G-L fractional derivative is particularly simple to implement with low computational cost. Its numerical calculation methods have been discussed in numerous published papers (e.g., [38]). Therefore, in this model, we replace integer-order derivatives with G-L defined fractional derivatives, constructing our diffusion coefficient.

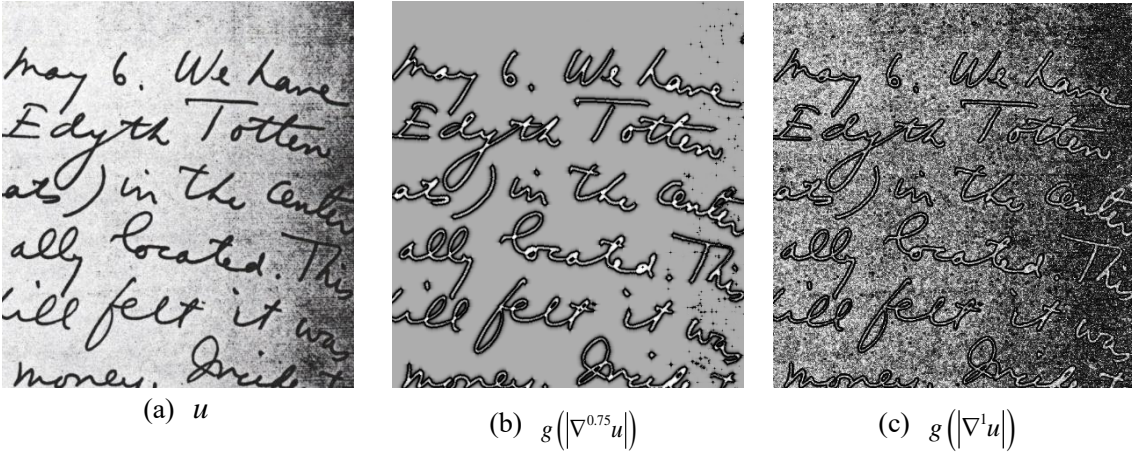

(a) $u$      (b) $g\left(\left|\nabla^{0.75}u\right|\right)$      (c) $g\left(\left|\nabla^1 u\right|\right)$

Fig. 2 Illustration of diffusion coefficients based on fractional and integer-order derivatives

To clarify the reason for using fractional-order derivatives in the diffusion coefficient of our model (instead of traditional integer-order derivatives), we illustrate in Fig.2(b)(c) the effect of $g\left(\left|\nabla^1 u\right|\right)$ and $g\left(\left|\nabla^{0.75}u\right|\right)$ on the degraded text image Fig.2(a).

Here, $\left|\nabla^{0.75}u\right|$ is computed using the G-L fractional derivative. Observing the text regions, we find that $g\left(\left|\nabla^{0.75}u\right|\right) \approx g\left(\left|\nabla u\right|\right)$, while in degraded background areas, $g\left(\left|\nabla^{0.75}u\right|\right) \gg g\left(\left|\nabla u\right|\right)$. Thus, during the diffusion process, compared to $g\left(\left|\nabla u\right|\right)$, $g\left(\left|\nabla^{0.75}u\right|\right)$ preserves text information while ensuring smoothness in degraded regions.

**(C) Binarization Term**

The term $B(s,u)$ in the partial differential equation system (19) is a binary term, defined as follows:

$$B(s,u) = a_{23}\omega(\mathrm{x})u(1-u)(u-c(\mathrm{x})) \\ +(1-a_{23})(1-\omega(\mathrm{x}))\mu(t)u(1-u)(u-s_{\min}) \\ +a_{24}u(1-u)(u-M(\mathrm{x})) \qquad (23)$$

where $M(\mathrm{x}) = \max\limits_{\mathrm{y}\in\delta_r(\mathrm{x})} u(\mathrm{y})$, and $\delta_r(\mathrm{x})$ is the local neighborhood centered at $\mathrm{x}$ with radius $r > 0$. The coefficients $a_{23}, a_{24}$ are weights for different terms. The functions $\omega(\mathrm{x})$ and $c(\mathrm{x})$ are still defined based on (6) and (7) as weight and threshold functions. The parameter $s_{\min}$ represents the minimum intensity value in the input image domain. In our model experiments, we set $\mu(t) = 1 - e^{-t/20}$.

To analyze the effect of the binarization term $B(s,u)$, we first consider the following initial value problem:

$$\frac{\partial u}{\partial t} = u(1-u)(u-c(\mathrm{x})),\ u(\mathrm{x},0) = s(\mathrm{x}),\ \mathrm{x} \in \Omega \qquad (24)$$

**Theorem 1** Let $u(\mathrm{x},0) = s(\mathrm{x})$ be the input image function, satisfying $0 \leq u(\mathrm{x},0) \leq 1$. The solution $u(\mathrm{x},t)$ of equation (24) has the following properties:

(i) if $0 \leq u(\mathrm{x},0) < c(\mathrm{x})$, then $\lim\limits_{t\to+\infty} u(\mathrm{x},t) = 0$,

(ii) if $u(\mathrm{x},0) = c(\mathrm{x})$, then $\lim\limits_{t\to+\infty} u(\mathrm{x},t) = c(\mathrm{x})$,

(iii) if $c(\mathrm{x}) < u(\mathrm{x},0) \leq 1$, then $\lim\limits_{t\to+\infty} u(\mathrm{x},t) = 1$.

**Proof**: This conclusion is easy to prove. (ii) is obvious, and we provide the proof for (i) here. (iii) can be derived similarly. Define $f_1(u) = u(1-u)(u-c(\mathrm{x}))$, since $f_1(0) = 0$ and $f_1^{'}(0) = -c(\mathrm{x}) \leq 0$, we see that $u = 0$ is a stable equilibrium point of (24). Constructing the function

$$U(t,u) = u^2 \tag{25}$$

which satisfies positivity, asymptotic boundedness, and three conditions for continuity and differentiability.

Following the evolution of function U in (25), the partial derivative with respect to time $t$ is given by:

$$U'_{(e)}(t,u) \triangleq \frac{\partial U}{\partial t}(t,u) + \frac{\partial U}{\partial u}(t,u)\frac{\partial U}{\partial t} \\ = -2u^2(u-1)(u-c(\mathrm{x})) < 0, 0 \leq u(\mathrm{x},0) < c(\mathrm{x}) \tag{26}$$

Thus, the equilibrium point $u = 0$ is asymptotically stable. Then, we obtain: $\lim_{t \to +\infty} u(\mathrm{x},t) = 0, 0 \leq u(\mathrm{x},0) < c(\mathrm{x})$, completing the proof.

**Remark 2** By Theorem 1, the solution $u(\mathrm{x},t)$ of the following problem:

$$\frac{\partial u}{\partial t} = u(1-u)(u-s_{\min}), \ u(\mathrm{x},0) = s(\mathrm{x}), \ \mathrm{x} \in \Omega \tag{27}$$

satisfies:

·if $u(\mathrm{x},0) \neq s_{\min}$, then $\lim_{t \to +\infty} u(\mathrm{x},t) = 1$,

·If $u(\mathrm{x},0) = s_{\min}$, then $\lim_{t \to +\infty} u(\mathrm{x},t) = s_{\min}$.

**Remark 3** Consider the solution $u(\mathrm{x},t)$ of the following problem:

$$\frac{\partial u}{\partial t} = u(1-u)(u-M(\mathrm{x})), \ u(\mathrm{x},0) = s(\mathrm{x}), \ \mathrm{x} \in \Omega \tag{28}$$

Since $M(\mathrm{x}) = \max_{\mathrm{y} \in \delta_r(\mathrm{x})} u(\mathrm{y})$, it follows that $\partial u / \partial t \leq 0$, meaning that $u(\mathrm{x},t)$ continuously decreases during evolution.

Obviously, from (23), we know that our binarization term $B(s,u)$ is obtained by adding the local maximum force term $B_M(s,u) = u(1-u)(u-M(\mathrm{x}))$ to $B_0(s,u)$, We refer to $B_M(s,u)$ as the local maximum force term. According to [39], $B_0(s,u)$ has the following properties:

Based on Theorem 1, for "foreground" points x satisfying $0 \leq u(\mathrm{x},0) < c(\mathrm{x})$, the solution $u(\mathrm{x},t)$ of (24) approaches 0 (foreground) as $t \to \infty$. For "background" points x satisfying $u(\mathrm{x},0) > c(\mathrm{x})$, the solution $u(\mathrm{x},t)$ of (24)approaches 1 (background) as $t \to \infty$.

As shown in Figure 3(b), when applying only (24) for the evolution of degraded text images, the text pixels in region $\Omega_1$ can be correctly classified. However, many background pixels in region $\Omega_2$ are mistakenly classified as foreground.

To address the issue of background points in $\Omega_2$ being misclassified as text points, we utilize the coefficient $\mu(t)$ and the weighting function $\omega(x)$. By combining (24) and (25), we obtain the following equation:

$$\frac{\partial u}{\partial t} = a_{23}\omega(x)u(1-u)(u-c(x)) \\ +(1-a_{23})(1-\omega(x))\mu(t)u(1-u)(u-s_{\min}) \quad (29)$$

The initial condition is given by $u(x,0) = s(x)$), where $\mu(t) = 1 - e^{-t/20}$ is increasing and satisfies $\mu(0) = 0$, and $\lim_{t \to \infty} \mu(t) = 1$.

In (29), the coefficient $(1-a_{23})\mu(t)\omega(x)$ only plays a role in the later stages of iteration and takes relatively large values only in the region $\Omega_2$. Therefore, the newly added second term primarily affects the region $\Omega_2$ in the later stages of iteration. Since $u(1-u)(u-s_{\min}) \geq 0$, it can gradually render the points in $\Omega_2$ more similar to the background, thereby counteracting the misclassification effect introduced by the first term. Figure 3(c) presents the binarization results of model (29). Comparing with Figure 3(c), it is evident that the inclusion of $u(1-u)(u-s_{\min})$ and the coefficient in model (29) leads to better binarization results compared to Figure 3(a). At this stage, a large number of misclassified pixels from (24) have been corrected.

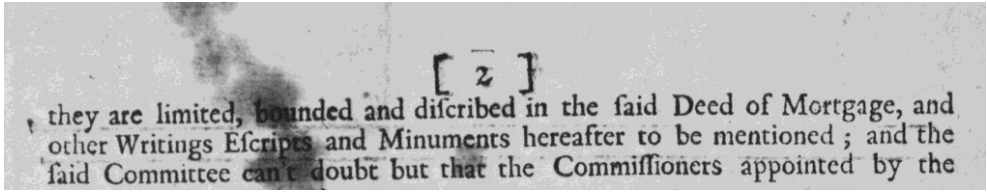

(a) origin u

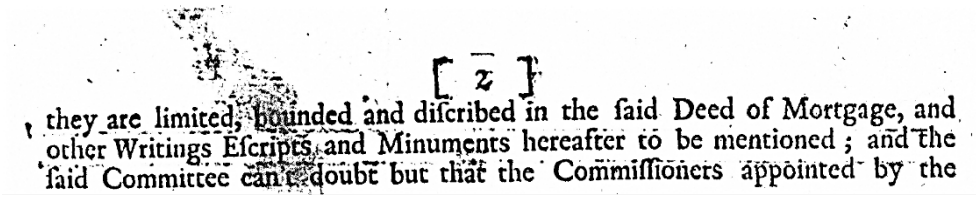

(b) image u by (24)

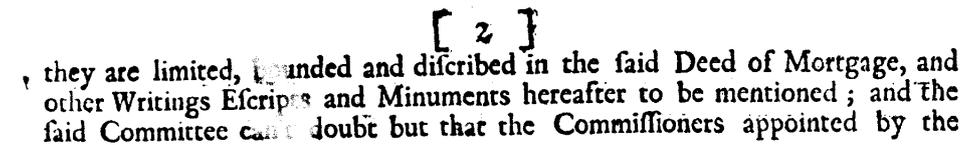

(c) image u by (29)

> they are limited, bounded and diſcribed in the ſaid Deed of Mortgage, and other Writings Eſcripts and Minuments hereafter to be mentioned ; and the ſaid Committee can't doubt but that the Commiſſioners appointed by the

(d) image u by (30)

Fig. 3 Evolution results of different equations

However, according to Remark 2, we know that $\partial u/\partial t = u(1-u)(u-s_{\min}) \geq 0$, While correcting the misclassification of background points, it may also cause some weak boundary text points to be misclassified, as shown in Figure 3(b). Therefore, to solve this problem, we introduce a local maximum force term $B_M(s,u)$ based on $B_0(s,u)$, leading to the final source term as follows:

$$\begin{aligned}\frac{\partial u}{\partial t} &= a_{23}\omega(\mathrm{x})u(1-u)(u-c(\mathrm{x}))\\&+(1-a_{23})(1-\omega(\mathrm{x}))\mu(t)u(1-u)(u-s_{\min})\\&+a_{24}u(1-u)(u-M(\mathrm{x}))\end{aligned} \quad (30)$$

where the initial condition is $u(\mathrm{x},0) = s(\mathrm{x})$. According to Remark 3, $B_M(s,u)$ satisfies $\partial u/\partial t = u(1-u)(u-M(\mathrm{x})) \leq 0$, which means that to some extent, it helps correctly classify weak boundary text. As shown in Fig.3(d), compared with Fig.3(c), we can see that the introduction of the local maximum force term enables (30) to better preserve weakly written text in the image during the rendering process.

### 3 Numerical algorithm

The proposed model (19) is formulated as a system of partial differential equations with Neumann boundary conditions. To numerically solve the proposed model, we extend the image at the boundary through mirror reflection. To numerically compute the integral (convolution) of the given function s(x) and weight function ω(x), we first provide the following explanation:

$$\int_{\mathcal{O}_\rho(\mathrm{x})} K_\rho(\mathrm{x}-\mathrm{y})h(\mathrm{y})d\mathrm{y} = \int_{\mathcal{R}^2} K_\rho(\mathrm{x}-\mathrm{y})h(\mathrm{y})d\mathrm{y} = (K_\rho * h)(\mathrm{x}) \quad (31)$$

where $h(\mathrm{y}) = s(\mathrm{y}), m_{\mathcal{F}}(\mathrm{y})s(\mathrm{y}), m_{\mathcal{B}}(\mathrm{y})s(\mathrm{y}), m_{\mathcal{F}}(\mathrm{y}), m_{\mathcal{B}}(\mathrm{y})$.

To compute $K_\rho * h$, we first generate a convolution kernel related to $K_\rho$. According to reference [45], we can derive that $n_\rho = 2[\rho/\sqrt{2}]+1$, where $[\cdot]$ denotes the ceiling function. The filter kernel's central position is at $(p_m, q_m) = ([\rho/\sqrt{2}]+1, [\rho/\sqrt{2}]+1)$. The convolution coefficient (still denoted as $K_\rho$) is computed as follows:

$$(K_\rho)_{p,q} = C\rho^{-2}\exp\left(-1/\left(1-\left((p-p_m)^2+(q-q_m)^2\right)/\rho^2\right)\right), \; 1\leq p,q\leq n_\rho \quad (32)$$

where $C$ is a normalization constant:

$$C = 1/\sum_{p,q=1}^{n_\rho} \rho^{-2} \exp\left(-1/\left(1-\left((x-x_m)^2+(y-y_m)^2\right)/\rho^2\right)\right) \quad (33)$$

which ensures $\sum_{p,q=1}^{n_\rho}(K_\rho)_{p,q} = 1$.

The proposed system of partial differential equations consists of two coupled equations. In numerical implementation, they are solved alternately to iteratively approximate the solution. Specifically, suppose that in the $n$ th iteration, there is an intermediate variable $b^n = b(\mathbf{x}, n\tau)$ and $u^n = u(\mathbf{x}, n\tau)$, with ($b^0 = 1$, $u^0 = s(\mathbf{x})$), where $\tau > 0$ is the time step.

First, we solve the equation:

$$\frac{\partial b}{\partial t} = a_{11} \cdot \Delta b + a_{12} \cdot u^n (s-b-u^n), \quad b(\mathbf{x},0)=b^n, \quad \frac{\partial b}{\partial \vec{n}} = 0 \quad (34)$$

by directly setting time $t = (n+1)\tau > 0$, we obtain $b^{n+1} = b(\mathbf{x}, (n+1)\tau)$.

Next, we solve:

$$\frac{\partial u}{\partial t} = a_{21} \cdot \mathrm{div}\left(g\left(|\nabla^\alpha u|\right)\nabla u\right) + a_{22} \cdot b^{n+1}\left(s - b^{n+1} - u\right) + N(s,u),$$
$$u(\mathbf{x},0) = u^n, \quad \frac{\partial u}{\partial \vec{n}} = 0 \quad (35)$$

until time $t = (n+1)\tau > 0$, and obtain $u^{n+1} = u(\mathbf{x}, (n+1)\tau)$.

For solving equations (34) and (35), we directly adopt an explicit finite difference scheme. Choosing a spatial step size $h = 1$ and a time step size $\tau > 0$, we uniformly denote $b_{i,j}^n = b(\mathbf{x}, n\tau)$ and $u_{i,j}^n = u(\mathbf{x}, n\tau)$, where $(i,j) = \mathbf{x}$ represents the pixel position coordinates.

In the numerical scheme, the second-order derivatives $b_{xx}$ and $b_{yy}$ at the $n$ th iteration are approximated by the following formulas:

$$\begin{aligned}(b_{xx})_{i,j}^n &\approx b_{i+1,j}^n + b_{i-1,j}^n - 2b_{i,j}^n \\ (b_{yy})_{i,j}^n &\approx b_{i,j+1}^n + b_{i,j-1}^n - 2b_{i,j}^n\end{aligned} \quad (36)$$

Thus, the numerical scheme for equation (34) can be expressed as:

$$\frac{b_{i,j}^{n+1} - b_{i,j}^n}{\tau} = a_{11}\left(b_{i+1,j}^n + b_{i-1,j}^n + b_{i,j+1}^n + b_{i,j-1}^n - 4b_{i,j}^n\right) + a_{11} \cdot u_{i,j}^n \left(s_{i,j} - b_{i,j}^n - u_{i,j}^n\right) \quad (37)$$

the diffusion term in equation (35) can be approximated using the central difference scheme as follows:

$$\frac{\partial}{\partial x}\left(g\left(|\nabla^\alpha u|\right)u_x\right)^n_{i,j} \approx (gu_x)^n_{i+\frac{1}{2},j} - (gu_x)^n_{i-\frac{1}{2},j}$$
$$\approx g^n_{i+\frac{1}{2},j}(u^n_{i+1,j} - u^n_{i,j}) - g^n_{i-\frac{1}{2},j}(u^n_{i,j} - u^n_{i-1,j}) \tag{38}$$

where $g^n_{i\pm\frac{1}{2},j}$ is computed as:

$$g^n_{i\pm\frac{1}{2},j} = \frac{1}{2}(g^n_{i,j} + g^n_{i\pm1,j}),\ g^n_{i,j} = g(|\nabla^\alpha u^n_{i,j}|) \tag{39}$$

combining (38) and (39), we obtain:

$$\frac{\partial}{\partial x}\left(g\left(|\nabla^\alpha u|\right)u_x\right)^n_{i,j} \approx g^n_{i+\frac{1}{2},j}(u^n_{i+1,j} - u^n_{i,j}) - g^n_{i-\frac{1}{2},j}(u^n_{i,j} - u^n_{i-1,j}),$$
$$\approx \frac{(g^n_{i,j} + g^n_{i+1,j})}{2}u^n_{i+1,j} - \frac{g^n_{i,j} + 2g^n_{i+1,j} + g^n_{i-1,j}}{2}u^n_{i,j}$$
$$+ \frac{(g^n_{i-1,j} + g^n_{i,j})}{2}u^n_{i-1,j} \tag{40}$$

Likewise, the following approximation can be obtained by exchanging $i$ and $j$ in (40):

$$\frac{\partial}{\partial y}\left(g\left(|\nabla^\alpha u|\right)u_y\right)^n_{i,j} \approx g^n_{i,j+\frac{1}{2}}(u^n_{i,j+1} - u^n_{i,j}) - g^n_{i,j-\frac{1}{2}}(u^n_{i,j} - u^n_{i,j-1}),$$
$$\approx \frac{(g^n_{i,j} + g^n_{i,j+1})}{2}u^n_{i,j+1} - \frac{g^n_{i,j} + 2g^n_{i,j+1} + g^n_{i,j-1}}{2}u^n_{i,j}$$
$$+ \frac{(g^n_{i,j-1} + g^n_{i,j})}{2}u^n_{i,j-1} \tag{41}$$

Finally, the discretization of (35) is given as follows:

$$\frac{u^{n+1}_{i,j} - u^n_{i,j}}{\tau} = a_{21}\left(g\left(|\nabla^\alpha u|\right)\nabla u\right)^n_{i,j} + a_{22}\cdot b^{n+1}_{i,j}\left(s_{i,j} - b^{n+1}_{i,j} - u^n_{i,j}\right) + B(s_{i,j}, u^n_{i,j}) \tag{42}$$

where

$$B(s_{i,j}, u^n_{i,j}) = a_{23}w_{i,j}\cdot u^n_{i,j}(1-u^n_{i,j})(u^n_{i,j} - c_{i,j})$$
$$+ (1-a_{23})(1-w_{i,j})f(t)u^n_{i,j}(1-u^n_{i,j})(u^n_{i,j} - s_{\min_{i,j}}) \tag{43}$$
$$+ a_{24}u^n_{i,j}(1-u^n_{i,j})(u^n_{i,j} - M_{i,j})$$

Let $u^*_{i,j}$ denote the limit image of the image sequences $\{u^n_{i,j}\}$ as $n \to \infty$. Then the binarized image $(s_B)_{i,j}$ are obtained by:

$$(s_B)_{i,j} = \begin{cases} 1, & u^*_{i,j} > 0.5 \\ 0, & u^*_{i,j} \leq 0.5 \end{cases} \tag{44}$$

**4 Experiments**

To verify the improvements of our model compared to the DH model and its adaptability to various degraded images, we conduct both qualitative and quantitative comparisons with the DH model and six binarization methods based on partial differential equations. Performance evaluation is conducted based on four metrics: FM, Fps, PSNR, and DRD. Higher FM, Fps, and PSNR values, along with lower DRD values, indicate better binarization performance.

Next, due to space limitations, we selected nine representative test images (see Fig. 4) to demonstrate the superiority of our model's results. These images represent various types of degradation.

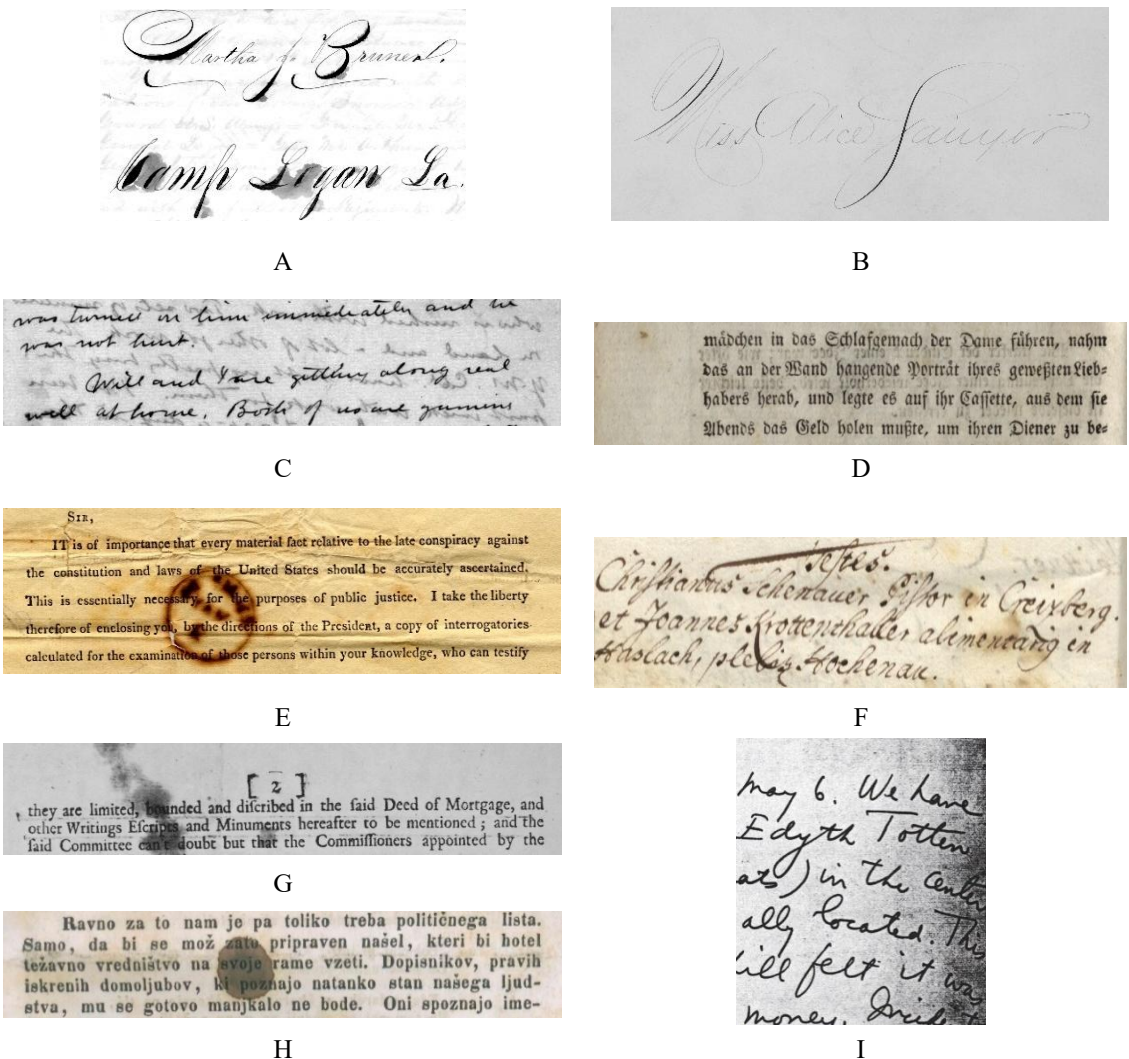

Fig. 4 Nine typical degraded document test images sourced from the DIBCO dataset

For blurred handwritten text images with water stains, the proposed model achieves excellent binarization results.

As shown in Figure 5, which presents the binarization result of Image A from Figure 4, it is visually evident that although the other seven models also achieve good binarization performance for this water-stained and blurred text image, our model better preserves the integrity of the text while minimizing the impact of water stain interference.

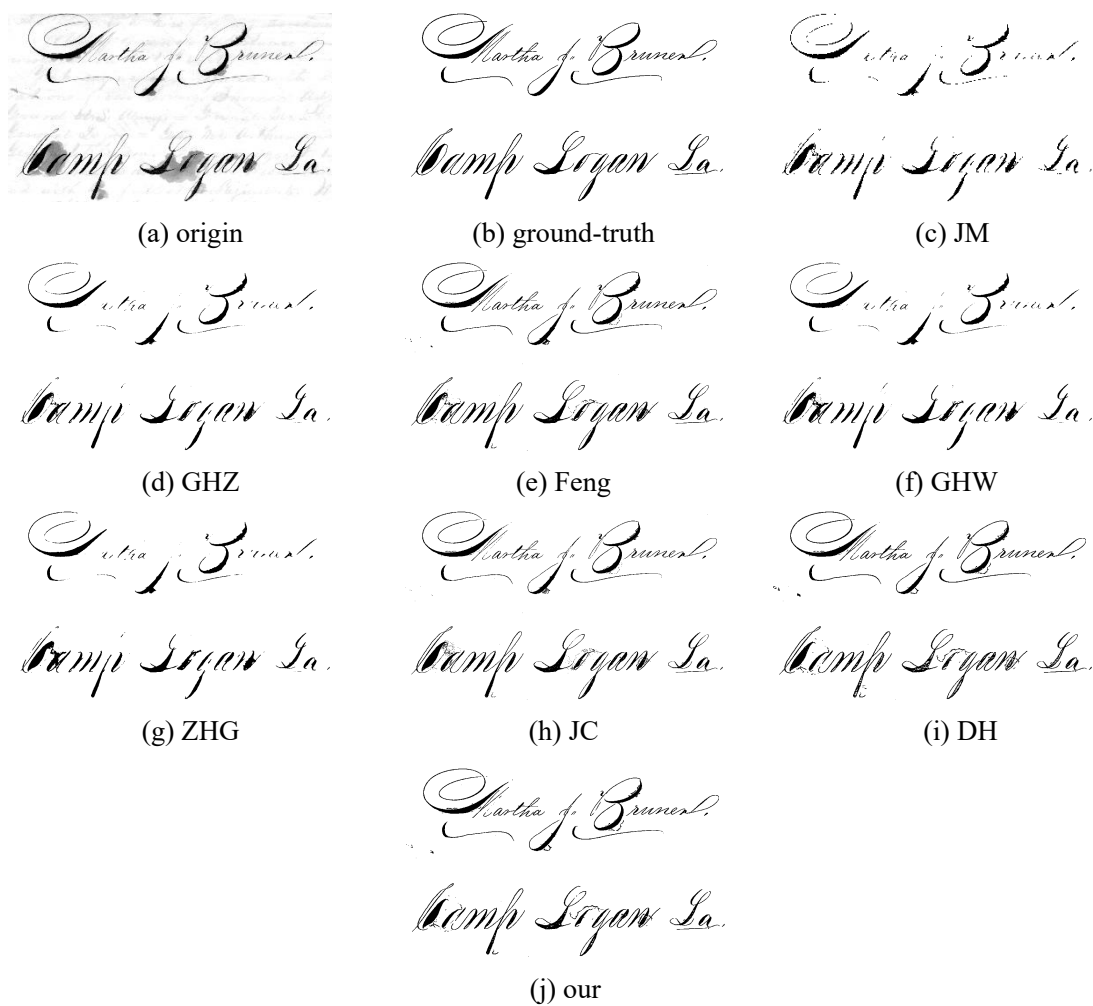

Fig. 5 Binarization results by eight methods for figure A

Figure 5 compares the binarization results of seven partial differential equation (PDE)-based models related to this work with the binarization result of our proposed model. As observed in the figure, the binarization results of JM, GHZ, GHW, and ZHG fail to accurately segment fine-stroke text. The binarization results of Feng and JC contain a significant amount of background noise. Although the DH model effectively addresses these issues, it introduces hollow strokes in certain areas. In contrast, our proposed model achieves a satisfactory binarization result by preserving fine strokes while effectively removing noise.

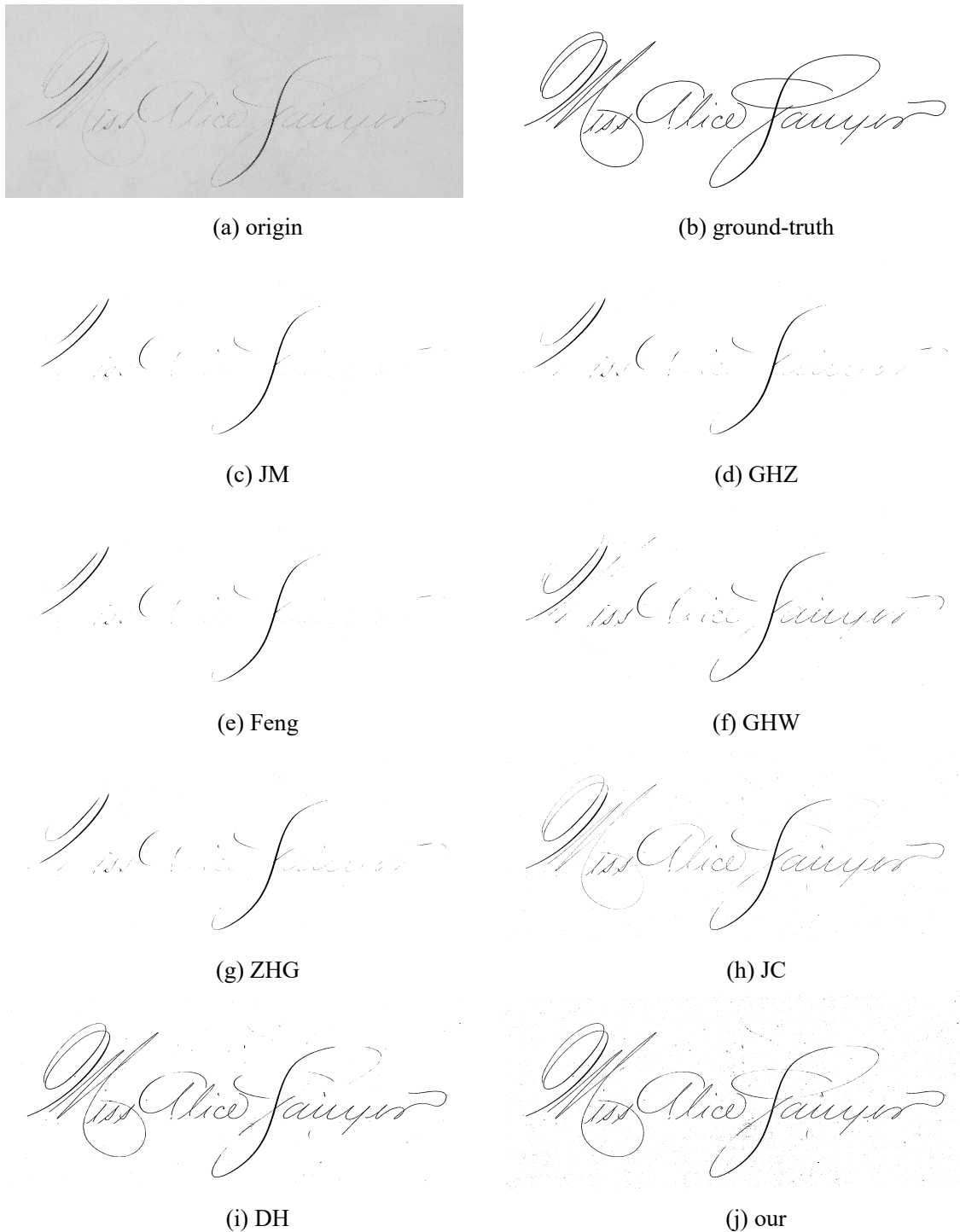

Fig. 6 Binarization results by eight methods for figure B

Figure 6 presents the binarization results of eight models applied to Image B from Figure 4. Image B is selected from the DIBCO 2013 dataset and contains a complex test image with uneven contrast and blurring. The goal of the binarization process is to address the blurring issue while preserving textual information. As shown in Figure 6, the binarization results of the JM, GHZ, Feng, and ZHG models exhibit significant text information loss. In contrast, the JC model, DH model, and our proposed model perform better in preserving most of the text information. Notably, our model demonstrates the best performance in maintaining text integrity, which can be specifically verified by the completeness of the strokes in the letter "f".

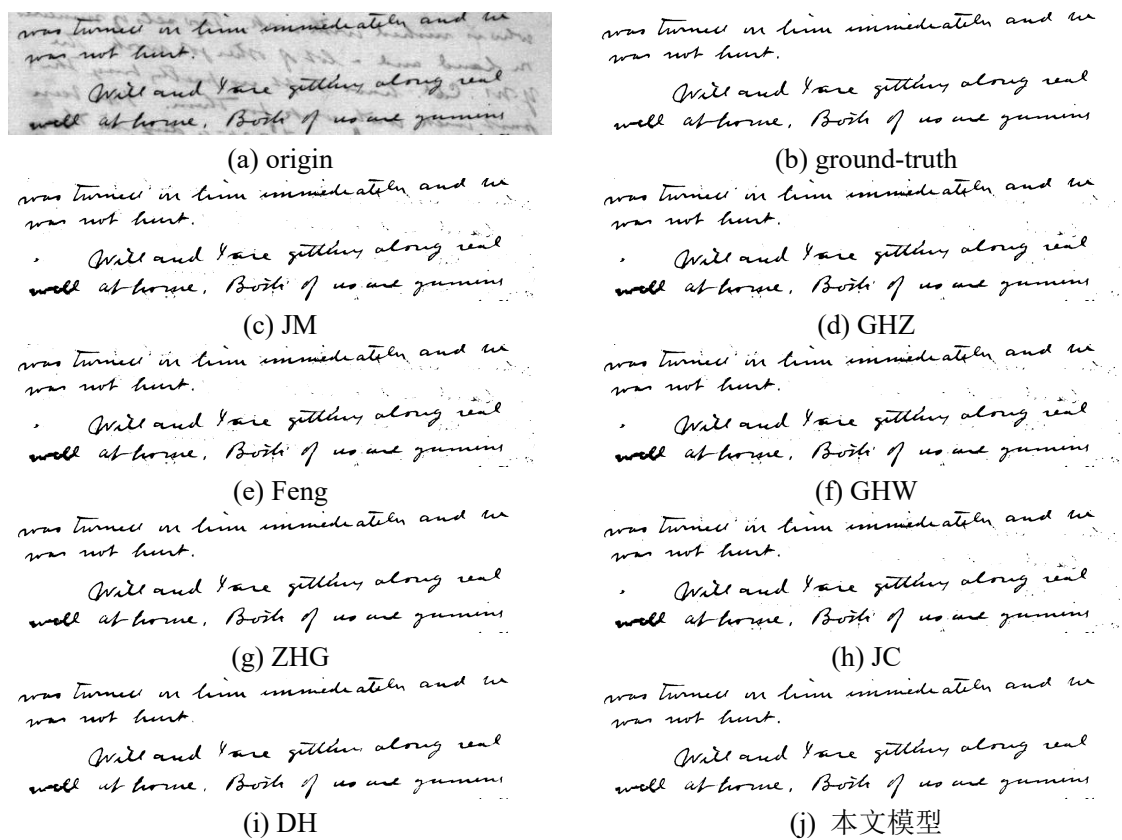

Fig. 7 Binarization results by eight methods for figure C

Figure 7 presents the binarization results of eight models applied to Image C from Figure 4. Image C is selected from the DIBCO 2010 dataset and contains a complex test image with degradation issues such as bleed-through and water stains. Clearly, our proposed model effectively addresses the bleed-through degradation problem while also accurately extracting text from the water-stained region in the lower-left corner of the image. Among the other models, all except the ZHG and DH models exhibit significant noise, indicating that their binarization results are still affected by bleed-through, leading to background pixels being misclassified as text. In comparison, the ZHG and DH models perform less effectively in handling water stain degradation, whereas our proposed model achieves the best performance in this aspect.

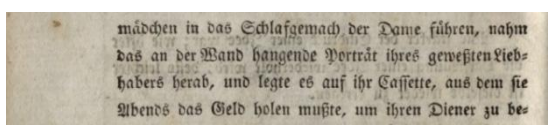

(a) origin

mädchen in das Schlafgemach der Dame führen, nahm
das an der Wand hangende Porträt ihres gewesten Lieb=
habers herab, und legte es auf ihr Cassette, aus dem sie
Abends das Geld holen mußte, um ihren Diener zu be=

(b) ground-truth

mädchen in das Schlafgemach der Dame führen, nahm
das an der Wand hangende Porträt ihres gewesten Lieb=
habers herab, und legte es auf ihr Cassette, aus dem sie
Abends das Geld holen mußte, um ihren Diener zu be=

(c) JM

mädchen in das Schlafgemach der Dame führen, nahm
das an der Wand hangende Porträt ihres gewesten Lieb=
habers herab, und legte es auf ihr Cassette, aus dem sie
Abends das Geld holen mußte, um ihren Diener zu be=

(d) GHZ

mädchen in das Schlafgemach der Dame führen, nahm
das an der Wand hangende Porträt ihres gewesten Lieb=
habers herab, und legte es auf ihr Cassette, aus dem sie
Abends das Geld holen mußte, um ihren Diener zu be=

(e) Feng

mädchen in das Schlafgemach der Dame führen, nahm
das an der Wand hangende Porträt ihres gewesten Lieb=
habers herab, und legte es auf ihr Cassette, aus dem sie
Abends das Geld holen mußte, um ihren Diener zu be=

(f) GHW

mädchen in das Schlafgemach der Dame führen, nahm
das an der Wand hangende Porträt ihres gewesten Lieb=
habers herab, und legte es auf ihr Cassette, aus dem sie
Abends das Geld holen mußte, um ihren Diener zu be=

(g) ZHG

mädchen in das Schlafgemach der Dame führen, nahm
das an der Wand hangende Porträt ihres gewesten Lieb=
habers herab, und legte es auf ihr Cassette, aus dem sie
Abends das Geld holen mußte, um ihren Diener zu be=

(h) JC

mädchen in das Schlafgemach der Dame führen, nahm
das an der Wand hangende Porträt ihres gewesten Lieb=
habers herab, und legte es auf ihr Cassette, aus dem sie
Abends das Geld holen mußte, um ihren Diener zu be=

(i) DH

(j) our

Fig.8 Binarization results by eight methods for figure D

Figure 8 presents the binarization results of eight models applied to Image D from Figure 4. Image D contains multiple types of degradation, including bleed-through, uneven illumination, and a complex background. As shown in Figure 8(j), our proposed model successfully preserves the complete text information in its binarization result. However, some noise remains, especially within the text itself. Among the other models, all except the ZHG and DH models are more severely affected by noise and misclassify background interference pixels in the upper-left corner as text, which are retained in the binarized result. The main drawback of the ZHG and DH models is their susceptibility to uneven illumination, which results in the text in the lower-right corner appearing thicker than in the reference image. In practical applications, this can lead to text misclassification, affecting subsequent text extraction and processing. To objectively evaluate the strengths and weaknesses of each model, we further assessed the algorithms using four evaluation metrics. The specific results are shown in Table 1.

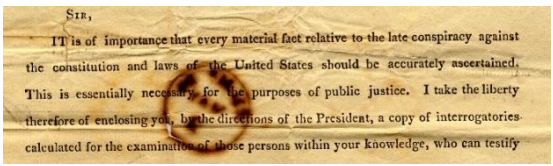

(a) origin

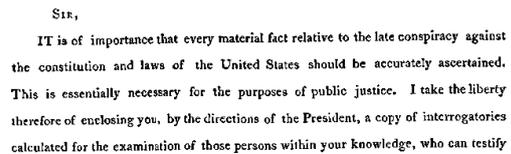

(b) ground-truth

(c) JM (d) GHZ

(e) Feng (f) GHW

(g) ZHG (h) JC

(i) DH (j) our

Fig. 9 Binarization results by eight methods for figure E

Figure 9 presents the binarization results of eight models applied to Image E from Figure 4. Image E contains prominent stamp stains, where the stamp not only covers the background but also obscures parts of the foreground text. The degraded regions exhibit strong grayscale differences compared to the non-degraded areas, posing a significant challenge for binarization. From the binarization results of the first six models, it is evident that many methods fail to recognize that the stamp is actually part of the background. In contrast, the DH model and our proposed model make significant progress in this regard, accurately identifying most of the pixels within the stamp as background pixels. However, some pixels within the stamp still belong to the foreground. A detailed comparison between Figure 9(i) and Figure 9(j) (such as the letter "y" in the fourth row and the phrase "examination of" in the fifth row) reveals that our proposed model has a clear advantage in preserving text information.

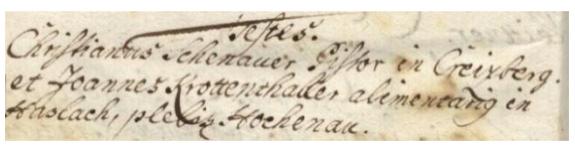 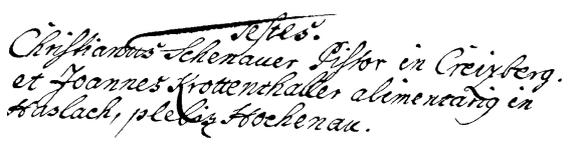

(a) origin (b) ground-truth

(c) JM

(d) GHZ

(e) Feng

(f) GHW

(g) ZHG

(h) JC

(i) DH

(j) our

Fig. 10 Binarization results by eight methods for figure F

Figure 10 visually presents the binarization results of eight models applied to Image F from Fig. 4. Image F is a handwritten test image with a complex background, uneven illumination, and varying stroke thicknesses. The goal of binarization for Image F is to preserve the integrity of the text while accurately distinguishing the foreground. The first few models share common characteristics: they effectively reduce the interference of the complex background and extract some text but perform poorly in maintaining text integrity. Specifically, many characters exhibit broken strokes (e.g., the first letter "C" in the image).In contrast, our proposed model addresses this issue more effectively. Although it is not the best among the eight models in terms of noise handling, it performs well in preserving text integrity. Further evaluation results of this test image can be found in Table 1.

(a) origin

(b) ground-truth

(c) JM

(d) GHZ

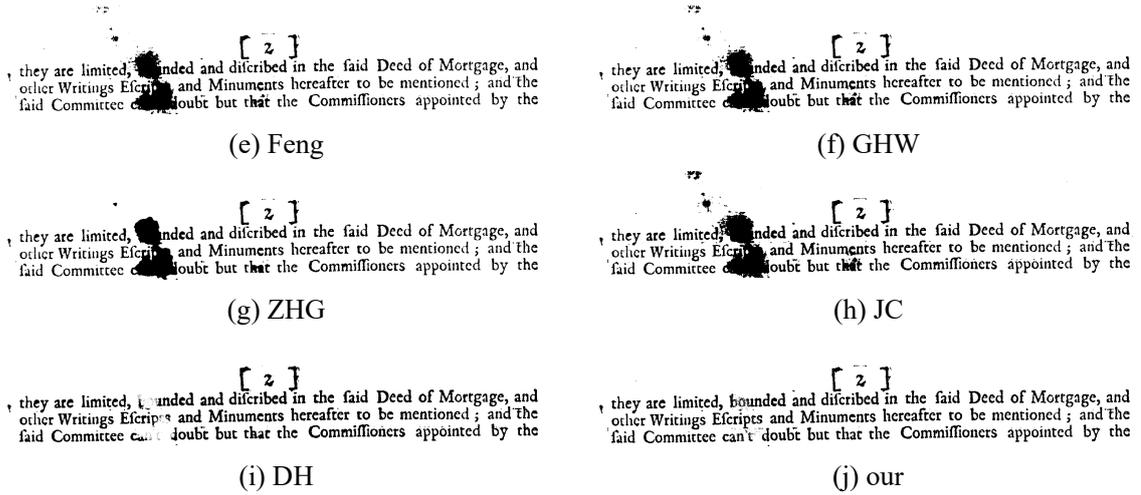

(e) Feng  (f) GHW

(g) ZHG  (h) JC

(i) DH  (j) our

Fig.11 Binarization results by eight methods for figure G

Figure 11 visually presents another grayscale image example similar to Image E from Figure 4. The original image, Image G from Figure 4, contains large areas of dark gray stains that cover both the text and background. The goal of binarization for this image is to accurately distinguish the pixels within the stains that should belong to the background while preserving the text in the stained regions as part of the foreground. For the first six PDE-based models, this poses a significant challenge. The DH model can remove the stains but almost entirely fails to retain the text covered by them. In contrast, our proposed model effectively addresses this issue.

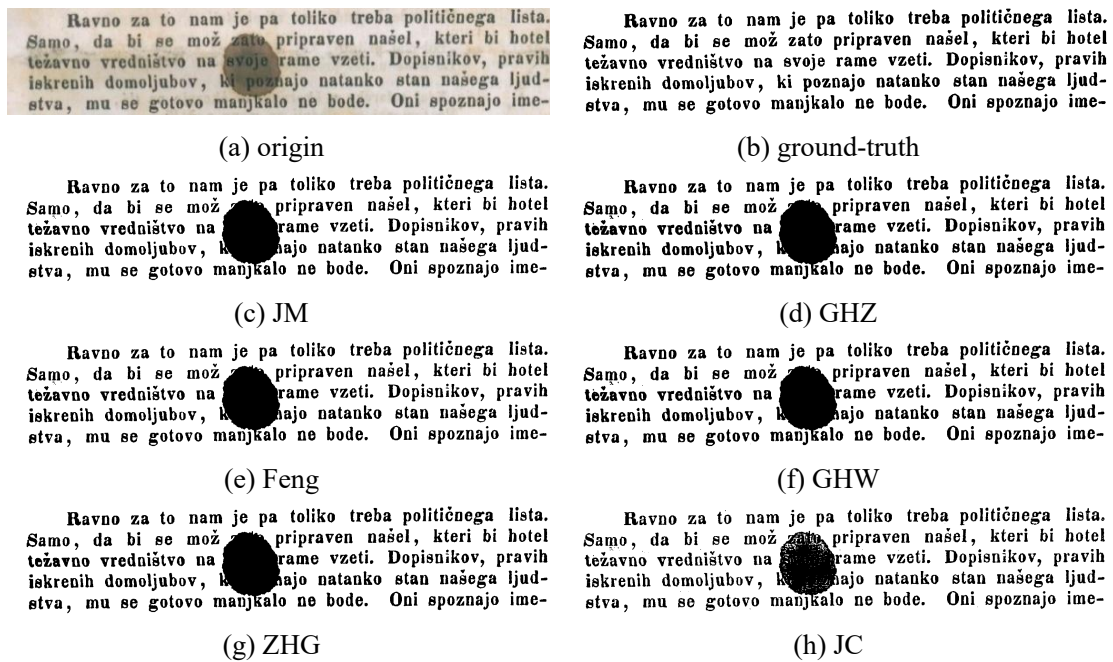

(a) origin  (b) ground-truth

(c) JM  (d) GHZ

(e) Feng  (f) GHW

(g) ZHG  (h) JC

          (i) DH                                       (j) our

Fig.12 Binarization results by eight methods for figure H

    Figure 12 presents the binarization results of eight models applied to Image H from Figure 4. Image H, selected from the DIBCO 2013 dataset, is a stained color-printed image. A comparison between Fig.12(i) and Fig.12(j) shows that although our proposed model exhibits some misclassification of pixels along the stain edges, the DH model also suffers from misclassification at the stain edges and significantly loses text information within the stained areas. In contrast, our proposed model better preserves the integrity of the text information.

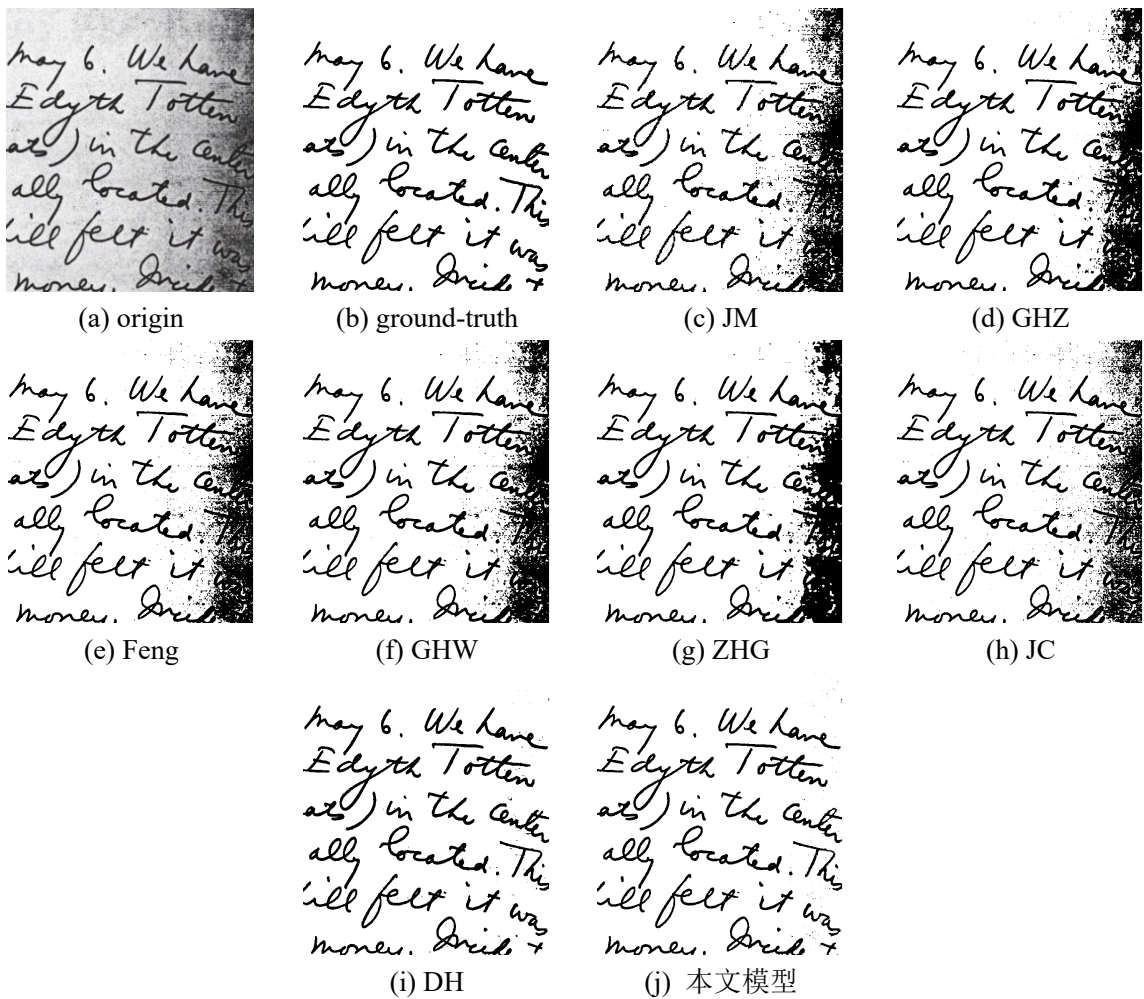

(a) origin     (b) ground-truth     (c) JM     (d) GHZ

(e) Feng     (f) GHW     (g) ZHG     (h) JC

(i) DH     (j) 本文模型

Fig.13 Binarization results by eight methods for figure I

    Figure 13 presents the binarization results of eight models applied to Image I from Fig. 4. Due to uneven illumination, the background on the right side of the image is prone to being

misclassified as the foreground, ultimately being retained as text in the binarized result. Additionally, the nature of binarized images makes it difficult to recognize text information in the affected areas. As shown in Fig.13, the first five models fail to resolve this issue. Although the JC model can recognize this part of the text, it still misclassifies most of the dark gray shadowed areas as text. The DH model and our proposed model perform better in handling this problem. A detailed comparison between Fig. 13(i) and Fig. 13(j) reveals that the DH model still exhibits some text loss in its binarization result (e.g., the letter "t" in the last word of the third row). In contrast, our proposed model successfully preserves complete text information and demonstrates superior noise handling. The objective evaluation metrics for this test can be found in Table 1.

Table 1. Evaluation measures of the eight methods for the nine degraded document images in Fig. 4

| Figure | measure | JM | GHZ | Feng | GHW | ZHG | JC | DH | Ours |
|---|---|---|---|---|---|---|---|---|---|
| (A) | FM(%) | 82.49 | 82.54 | 81.29 | 82.41 | 82.02 | 91.25 | 92.50 | **93.72** |
| | $F_{PS}$(%) | 74.15 | 4.25 | 87.63 | 74.70 | 72.93 | **94.61** | 93.38 | 94.37 |
| | PSNR | 17.50 | 17.51 | 15.42 | 17.39 | 17.30 | 20.16 | 20.75 | **21.51** |
| | DRD | 8.49 | 8.47 | 7.24 | 8.81 | 9.08 | 4.60 | 4.44 | **3.48** |
| (B) | FM(%) | 32.52 | 37.72 | 32.65 | 52.00 | 37.74 | 60.93 | 76.43 | **80.55** |
| | $F_{PS}$(%) | 35.09 | 41.12 | 35.30 | 60.54 | 42.63 | 78.30 | **89.76** | 88.73 |
| | PSNR | 17.33 | 17.53 | 17.33 | 18.26 | 17.54 | 18.85 | 20.51 | **20.76** |
| | DRD | 8.90 | 8.47 | 8.89 | 7.04 | 8.45 | 6.05 | **4.13** | 4.46 |
| (C) | FM(%) | 90.91 | 91.28 | 90.87 | 91.01 | **94.02** | 91.25 | 91.18 | 92.66 |
| | $F_{PS}$(%) | 91.18 | 91.44 | 91.16 | 90.98 | 95.11 | 91.46 | 93.27 | **96.12** |
| | PSNR | 19.57 | 19.78 | 19.55 | 19.60 | **21.58** | 19.75 | 19.97 | 20.83 |
| | DRD | 3.17 | 2.81 | 3.18 | 3.12 | **1.62** | 2.94 | 2.45 | 2.07 |
| (D) | FM(%) | 91.73 | 91.77 | 91.73 | 92.02 | **92.77** | 91.84 | 92.09 | 92.21 |
| | $F_{PS}$(%) | 95.19 | 95.34 | 95.21 | 94.82 | 95.36 | 94.80 | 91.90 | **95.46** |
| | PSNR | 17.07 | 17.09 | 17.07 | 17.17 | **17.66** | 17.09 | 17.04 | 17.34 |
| | DRD | 2.32 | 2.30 | 2.31 | 2.26 | **1.85** | 2.30 | 2.13 | 2.08 |
| (E) | FM(%) | 51.24 | 51.24 | 51.17 | 51.26 | 51.3 | 73.74 | 89.90 | **90.79** |
| | $F_{PS}$(%) | 49.57 | 49.54 | 49.51 | 49.28 | 49.52 | 74.52 | 97.67 | **97.87** |
|     | PSNR    | 6.71  | 6.70  | 6.70  | 6.60  | 6.70  | 12.62 | 17.94 | **18.29** |
|     | DRD     | 56.99 | 57.19 | 57.09 | 58.77 | 57.24 | 12.72 | 2.53  | **2.30**  |
| (F) | FM(%)   | 84.98 | 89.01 | 85.00 | 90.00 | 90.95 | **90.91** | 89.32 | 90.22 |
|     | $F_{PS}$(%) | 84.77 | 85.06 | 84.81 | 86.60 | 87.82 | **92.57** | 85.73 | 89.89 |
|     | PSNR    | 14.97 | 15.73 | 14.98 | 16.36 | 16.70 | **16.84** | 15.84 | 16.18 |
|     | DRD     | 2.98  | 2.70  | 2.97  | 2.21  | 2.12  | **2.14**  | 2.51  | 2.81  |
| (G) | FM(%)   | 82.75 | 82.79 | 85.78 | 83.52 | 83.62 | 83.27 | 92.57 | **93.92** |
|     | $F_{PS}$(%) | 87.07 | 87.06 | 87.17 | 86.15 | 87.43 | 84.14 | 93.09 | **96.24** |
|     | PSNR    | 14.35 | 14.37 | 12.55 | 14.42 | 14.58 | 14.17 | 18.04 | **19.00** |
|     | DRD     | 7.22  | 7.17  | 8.85  | 7.36  | 6.84  | 8.24  | 2.65  | **2.03** |
| (H) | FM(%)   | 85.78 | 87.20 | 82.76 | 87.02 | 87.06 | 85.15 | 93.87 | **94.30** |
|     | $F_{PS}$(%) | 87.17 | 86.24 | 87.10 | 86.89 | 87.28 | 88.09 | 95.92 | **96.62** |
|     | PSNR    | 12.55 | 12.83 | 14.35 | 12.84 | 12.89 | 12.48 | 16.49 | **16.80** |
|     | DRD     | 8.85  | 8.50  | 7.20  | 8.46  | 8.38  | 8.82  | 2.83  | **2.69** |
| (I) | FM(%)   | 70.84 | 70.91 | 70.85 | 69.99 | 70.74 | 72.01 | 93.58 | **94.06** |
|     | $F_{PS}$(%) | 69.76 | 69.75 | 69.78 | 68.75 | 69.52 | 71.62 | 96.40 | **98.33** |
|     | PSNR    | 10.05 | 10.06 | 10.06 | 9.87  | 10.03 | 10.37 | 18.10 | **18.39** |
|     | DRD     | 22.66 | 22.50 | 22.64 | 23.67 | 22.72 | 20.94 | 2.39  | **2.17** |

Table 1 presents the numerical values of four evaluation metrics (FM, Fps, PSNR, and DRD) corresponding to the binarization results of Images A–I from Figure 4 for the eight models. From the table, it is evident that our proposed model demonstrates significantly superior performance compared to the other eight models. Although for Images C and D, our model performs slightly worse than the ZHG model in terms of the evaluation metrics, the difference is minimal. Moreover, the ZHG model mainly excels in processing text images with bleed-through degradation, such as Images C and D.

Therefore, from both subjective and objective perspectives, our proposed model has a notable advantage in handling these types of degraded images. In particular, it enhances binarization performance based on the effectiveness of the DH model while better preserving weak text edges.

To further demonstrate the superior generalizability of our model in processing various degraded text images, Table 2 provides the statistical values of the four objective metrics across 86 images from the selected dataset for all eight models.

Table 2. Average values of evaluation measures of eight methods on seven DIBCO datasets

| measure | model | 2009 | 2010 | 2011 | 2012 | 2013 | 2014 | 2016 | ALL |
|---|---|---|---|---|---|---|---|---|---|
| FM(%)(↑) | JM | 75.71 | 67.62 | 77.90 | 81.57 | 77.72 | 74.23 | 83.25 | 77.21 |
| | GHZ | 83.31 | 86.75 | 83.82 | 86.40 | 82.35 | 92.30 | 88.51 | 85.78 |
| | Feng | 78.41 | 61.11 | 76.80 | 72.11 | 77.71 | 74.03 | 83.26 | 75.00 |
| | GHW | 81.85 | 88.56 | 84.26 | 87.48 | 83.88 | 92.3 | 88.31 | 86.34 |
| | ZHG | 84.06 | 85.70 | 84.40 | 86.4 | 82.85 | 91.11 | 88.04 | 85.75 |
| | JC | 83.16 | 79.04 | 70.32 | 82.79 | 83.96 | 82.02 | 84.31 | 80.38 |
| | DH | 91.97 | 88.74 | 90.00 | 91.12 | 91.85 | **94.57** | 90.55 | 91.20 |
| | Ours | **92.22** | **90.26** | **90.65** | **91.82** | **92.19** | 94.37 | **91.76** | **91.83** |
| $F_{PS}$(%)(↑) | JM | 76.46 | 68.78 | 74.60 | 81.45 | 82.34 | 81.29 | 87.98 | 79.03 |
| | GHZ | 84.91 | 89.74 | 87.32 | 89.00 | 85.16 | 94.86 | 90.46 | 88.43 |
| | Feng | 81.51 | 70.15 | 81.93 | 77.93 | 82.35 | 81.12 | 88.03 | 80.55 |
| | GHW | 83.22 | 93.50 | 88.16 | 90.16 | 87.25 | 96.62 | 91.27 | 89.71 |
| | ZHG | 86.00 | 90.61 | 88.43 | 89.37 | 86.47 | 95.24 | 90.89 | 89.27 |
| | JC | 85.23 | 89.36 | 73.61 | 89.77 | 90.27 | 87.40 | 89.51 | 85.98 |
| | DH | 94.29 | 92.31 | 93.61 | 93.70 | 95.13 | **97.28** | 91.66 | 94.04 |
| | Ours | **95.02** | **93.71** | **94.80** | **94.70** | **95.80** | 97.02 | **92.99** | **94.91** |
| PSNR(↑) | JM | 13.74 | 14.66 | 14.68 | 16.26 | 16.54 | 15.41 | 17.66 | 15.60 |
| | GHZ | 16.64 | 17.93 | 16.5 | 17.86 | 17.37 | 19.17 | 18.42 | 17.60 |
| | Feng | 15.63 | 15.11 | 14.85 | 16.05 | 16.54 | 15.40 | 17.65 | 15.87 |
| | GHW | 16.30 | 18.56 | 16.5 | 18.04 | 17.54 | 19.05 | 18.41 | 17.68 |
| | ZHG | 16.96 | 17.96 | 16.69 | 17.96 | 17.81 | 18.53 | 18.58 | 17.72 |
| | JC | 16.24 | 16.59 | 12.91 | 17.18 | 17.27 | 16.74 | 17.71 | 16.24 |
| | DH | 18.65 | 18.70 | 18.38 | 19.73 | **20.15** | **20.22** | 18.94 | 19.27 |
| | Ours | **19.01** | **19.08** | **18.52** | **19.87** | 20.10 | 20.11 | **19.57** | **19.46** |
| DRD(↓) | JM | 7.46 | 8.02 | 6.28 | 5.98 | 9.20 | 7.44 | 5.32 | 7.14 |
| | GHZ | 9.40 | 3.62 | 6.55 | 4.67 | 8.09 | 2.37 | 4.13 | 5.75 |

| | | | | | | | | |
|---|---|---|---|---|---|---|---|---|
| Feng | 10.62 | 8.27 | 16.44 | 7.54 | 9.20 | 7.47 | 5.32 | 9.68 |
| GHW | 13.41 | 3.06 | 6.81 | 4.68 | 8.28 | 2.35 | 4.26 | 6.25 |
| ZHG | 8.70 | 3.60 | 6.35 | 4.45 | 7.63 | 2.73 | 4.00 | 5.54 |
| JC | 10.29 | 5.49 | 53.47 | 5.63 | 6.04 | 5.76 | 5.60 | 15.14 |
| DH | 2.73 | 3.20 | 2.94 | 2.99 | **2.35** | **1.63** | 3.41 | 2.74 |
| Ours | **2.61** | **2.72** | **2.84** | **2.74** | 2.40 | 1.74 | **3.07** | **2.60** |

From Table 2, it can be observed that our proposed model demonstrates an absolute advantage across all four evaluation metrics for the binarization results on the DIBCO 2009–DIBCO 2012 and DIBCO 2016 datasets. Although on the DIBCO 2014 dataset, the FM and Fps metrics are slightly lower than those of the DH model, and on the DIBCO 2013–2014 datasets, the PSNR and DRD metrics are slightly lower than those of the DH model, the differences between our model and the DH model are minimal.

**5 Conclusions**

This paper first provides a detailed introduction to the partial differential equation (PDE) models related to text binarization and the design principles of the DH model. Next, the derivation process of the additive modeling for text images is presented. Based on the DH model and image modeling, we propose a weakly coupled PDE model to simultaneously estimate the background and foreground components. In this system of equations, the first equation is used to estimate the background component, while the second equation estimates the foreground component. The final binarization result is obtained by applying a hard projection to the estimated foreground component.

In estimating the foreground component, a fractional-order derivative-based diffusion coefficient is designed to ensure that the nonlinear diffusion term effectively smooths degraded regions while preserving the text. Additionally, a local maximum force term is introduced in the source term, enabling the PDE system to better protect weak text information in the image. To validate the effectiveness of the proposed model, 86 degraded text images from the DIBCO benchmark dataset were used for evaluation. Comparisons were conducted against seven mainstream PDE-based binarization models. Experimental results demonstrate that the proposed model has significant advantages in processing degraded text images. Furthermore, quantitative analysis using multiple evaluation metrics, including FM, Fps, PSNR, and DRD, further confirms the superiority of the proposed model in binarization performance.

**References**

[1] W. Xiong, J. Xu, Z. Xiong, J. Wang, and M. Liu. Degraded historical document image binarization using local features and support vector machine (SVM)[J]. Optik, 2018, 164: 218-223.

[2] E.-T. Zemouri and Y. Chibani. Nonsubsampled contourlet transform and k-means clustering for degraded document image binarization[J]. Journal of Electronic Imaging, 2019, 28.

[3] P. Jana, S. Ghosh, R. Sarkar, and M. Nasipuri. A fuzzy c-means based approach towards efficient document image binarization[C]//2017 Ninth International Conference on Advances in Pattern Recognition. Bangalore: IEEE, 2017: 332-337.

[4] Q. N. Vo, S. H. Kim, H. J. Yang, and G. Lee. Binarization of degraded document images based on hierarchical deep supervised network[J]. Pattern Recognition, 2018, 74: 568-586.

[5] J. Calvo-Zaragoza and A.-J. Gallego. A selectional auto-encoder approach for document image binarization[J]. Pattern Recognition, 2019, 86: 37-47.

[6] S. He and L. Schomaker. Deepotsu: Document enhancement and binarization using iterative deep learning[J]. Pattern Recognition, 2019, 91: 379-390.

[7] S. He and L. Schomaker. Ct-net: Cascade t-shape deep fusion networks for document binarization[J]. Pattern Recognition, 2021, 118.

[8] S. Han, S. Ji, and J. Rhee. Diffusion-denoising process with gated u-net for high-quality document binarization[J]. Applied Sciences-Basel, 2023, 13.

[9] J. Zhao, C. Shi, F. Jia, Y. Wang, and B. Xiao. Document image binarization with cascaded generators of conditional generative adversarial networks[J]. Pattern Recognition, 2019, 96.

[10] S. K. Jemni, M. A. Souibgui, Y. Kessentini, and A. Fornes. Enhance to read better: A multi-task adversarial network for handwritten document image enhancement[J]. Pattern Recognition, 2022, 123.

[11] M. A. Souibgui and Y. Kessentini. De-gan: A conditional generative adversarial network for document enhancement[J]. IEEE Transactions on Pattern Analysis and Machine Intelligence, 2022, 44: 1180-1191.

[12] M. Yang and S. Xu. A novel degraded document binarization model through vision transformer network[J]. Information Fusion, 2023, 93: 159-173.

[13] N. OTSU. A threshold selection method from gray-level histograms[J]. IEEE Transactions on Systems, Man, and Cybernetics, 1979, 9(1): 62-66.

[14] J. M. Prewitt and M. L. Mendelsohn. The analysis of cell images[J]. Annals of the New York Academy of Sciences, 1966, 128: 1035-1053.

[15] W. Niblack. An introduction to digital image processing[M]. Birkeroed: Strandberg Publishing Company, 1986: 411-416.

[16] J. Sauvola and M. Pietikäinen. Adaptive document image binarization[J]. Pattern Recognition, 2000, 33: 225-236.

[17] C. Wolf, J. Jolion, and F. Chassaing. Text localization, enhancement and binarization in multimedia documents[C]//International Conference on Pattern Recognition. Quebec City: IEEE, 2002: 1037-1040.

[18] J. Bernsen. Dynamic thresholding of gray level image[C]//International Conference on Pattern Recognition. Berlin: IEEE, 1986: 1251-1255.

[19] B. Su, S. Lu, and C. L. Tan. Binarization of historical document images using the local maximum and minimum[C]//9th International Workshop on Document Analysis Systems. New York: Association for Computing Machinery, 2010: 159-166.

[20] B. Su, S. Lu, and C. L. Tan. Robust document image binarization technique for degraded document images[J]. IEEE Transactions on Image Processing, 2013, 22: 1408-1417.

[21] B. Gatos, I. Pratikakis, and S. Perantonis. Adaptive degraded document image binarization[J]. Pattern Recognition, 2006, 39: 317-327.

[22] N. R. Howe. Document binarization with automatic parameter tuning[J]. International Journal on Document Analysis and Recognition, 2013, 16: 247-258.

[23] R. DERICHE. Using canny's criteria to derive a recursively implemented optimal edge detector[J]. International Journal of Computer Vision, 1987, 1(2): 167-187.

[24] E. Ahmadi, Z. Azimifar, M. Shams, M. Famouri, and M. J. Shafiee. Document image binarization using a discriminative structural classifier[J]. Pattern Recognition Letters, 2015, 63: 36-42.

[25] F. Jia, C. Shi, K. He, C. Wang, and B. Xiao. Document image binarization using structural symmetry of strokes[C]//15th International Conference on Frontiers in Handwriting Recognition. Shenzhen: IEEE, 2016: 411-416.

[26] F. Jia, C. Shi, K. He, C. Wang, and B. Xiao. Degraded document image binarization using structural symmetry of strokes[J]. Pattern Recognition, 2018, 74: 225-240.

[27] Y. Wang, C. He. Binarization method based on evolution equation for document images produced by cameras [J]. Journal of Electronic Imaging, 2012, 21(2): 3030.

[28] B. A. Jacobs, E. Momoniat. A novel approach to text binarization via a diffusion-based model [J]. Applied Mathematics and Computation, 2013, 225: 446-460.

[29] B. Jacobs, E. Momoniat. A locally adaptive, diffusion based text binarization technique, Applied Mathematics and Computation, 269 (2015) 464-472.

[30] J. Guo, C. He, X. Zhang. Nonlinear edge-preserving diffusion with adaptive source for document images binarization[J]. Applied Mathematics and Computation, 2019, 351: 8-22.

[31] X. Zhang, C. He, J. Guo. Selective diffusion involving reaction for binarization of bleed-through document images[J]. Applied Mathematical Modelling, 2020, 81: 844-854.

[32] J. Guo, C. He, and Y. Wang. Fourth order indirect diffusion coupled with shock filter and source for text binarization[J]. Signal Processing, 2020, 171: 107478.

[33] S. Feng. A novel variational model for noise robust document image binarization [J]. Neurocomputing, 2019, 325: 288-302.

[34] D. Mumford, J. Shah. Optimal approximations by piecewise smooth functions and associated variational problems [J]. Communications on Pure and Applied Mathematics, 1989, 42(5): 577-685.

[35] L. Rudin, S. Osher, E. Fatemi. Nonlinear total variation based noise removal algorithms [J]. Physica D, 1992, 60(1-4): 259-268.

[36] Z. Du and C. He. Nonlinear diffusion system for simultaneous restoration and binarization of degraded document images[J]. Computers and Mathematics with Applications, 2024, 153: 237-249.

[37] P. PERONA and J. MALIK. Scale-space and edge detection using anisotropic diffusion[J]. IEEE Transactions on Pattern Analysis and Machine Intelligence, 1990, 12: 629-639.

[38] I. Podlubny. Fractional Differential Equations[M]. San Diego: Academic Press Inc., 1999.

[39] B. Jacobs and T. Celik. Unsupervised document image binarization using a system of nonlinear partial differential equations[J]. Applied Mathematics and Computation, 2022, 418: 126806.

[40] Z. Du and C. He. Nonlinear diffusion equation with a dynamic threshold-based source for text binarization[J]. Applied Mathematics and Computation.

[41] Jia F, Shi C, He K, Wang C, Xiao B. Degraded document image binarization using structural symmetry of strokes. Pattern Recognition, 2018; 74: 225–240.

[42] Khitas M, Ziet L, Bouguezel S. Improved degraded document image binarization using median flter for background estimation. Elektronika ir Elektrotechnika. 2018;24(3):82–7.

[43] Chaudhary P, Saini B. An effective and robust technique for the binarization of degraded document images[J]. Int J Res Eng Technol (IJRET), 2014, 3(06): 140.

[44] Gatos B, Pratikakis I, Perantonis SJ. Adaptive degraded document image binarization[J]. Pattern Recognit. 2006;39(3):317–27.

[45] X. Zhang and C. He. Robust double-weighted guided image filtering[J]. Signal Processing, 2022, 199: 108609.